\documentclass[final]{cvpr}

\usepackage{times}
\usepackage{epsfig}
\usepackage{graphicx}
\usepackage{amsmath}
\usepackage{amssymb}
\usepackage{epigraph}
\usepackage{comment}

\usepackage{blindtext}
\usepackage{caption}

\usepackage[pagebackref=true,breaklinks=true,colorlinks,bookmarks=false]{hyperref}

\def\cvprPaperID{****} 
\def\confYear{CVPR 2021}

\begin{document}

\title{A modular framework for object-based saccadic decisions in dynamic scenes}

\author{Nicolas Roth, \quad Pia Bideau, \quad Olaf Hellwich,\quad  Martin Rolfs, \quad Klaus Obermayer\\
Exzellenzcluster Science of Intelligence, Technische Universität Berlin\\
Marchstrasse 23, 10587 Berlin, Germany
}
\maketitle

\begin{abstract}
	Visually exploring the world around us is not a passive process. Instead, we actively explore the world and acquire visual information over time. Here, we present a new model for simulating human eye-movement behavior in dynamic real-world scenes. 
	We model this active 
	scene-exploration as a sequential decision making process.
	We adapt the popular drift-diffusion model (DDM) for perceptual decision making and extend it towards multiple options, defined by objects present in the scene.
	For each possible choice, the model integrates evidence over time and a decision (saccadic eye movement) is triggered as soon as evidence crosses a decision threshold. %
	Drawing this explicit connection between decision making and object-based scene perception is highly relevant in the context of active viewing, where decisions are made continuously 
	while interacting with an external environment.
	We validate our model with a carefully designed ablation study, and explore influences of our model parameters. A comparison on the VidCom dataset\footnote{Data: \url{http://ilab.usc.edu/vagba/dataset/VidCom/}} supports the plausibility of the proposed approach.
	

\end{abstract}

\section{Introduction}
Tackling the question of how humans capture visual information requires an understanding of the interplay between scene understanding \textit{(What do we see?)} and the generation of eye movements \textit{(What do we look at?)}.
In this work, we attempt to draw this connection and present a new model for object-based decision making in dynamic scenes.

\begin{figure*}
    \centering
    \includegraphics[width=.99\textwidth]{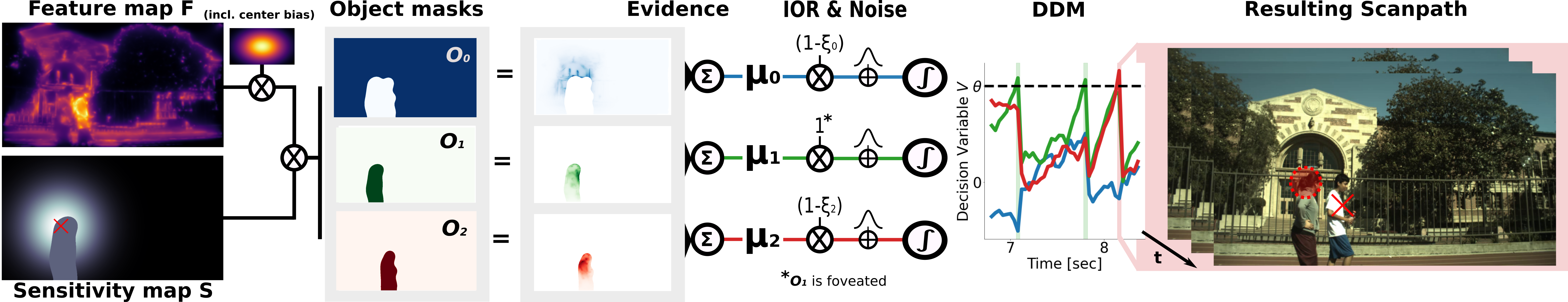}
    \caption{Calculation of the decision variable for each object, based on the visual input and current gaze location. The visual information $(F\cdot S)$ is accumulated for all object masks in the scene. A saccade is triggered towards an object $O_i$, as soon as the respective decision variable (blue for $O_0$, green for $O_1$, red for $O_2$) crosses the threshold (\href{https://github.com/rederoth/ObjectDDM-videos}{link to video of resulting scanpath}). The red cross denotes the current and the circle the previous one gaze position.}
    \label{fig:title}
\end{figure*}

A typical model for perceptual decision making is the drift-diffusion model (DDM) \cite{ratcliff2008diffusion}, which finds a wide range of applications in the area of behavioral decision making. 
The basic premise of DDMs is that evidence from a stimulus (drift) which is perturbed by noise (diffusion) is accumulated over time. A decision is reached once enough evidence has been sampled.
A similar approach, but limited to static inputs with constant accumulation rates, has been used by Tatler et al.~\cite{tatler2017latest} to model gaze control in static scenes.

Following the idea of a sequential decision making process, a saccade towards an object (or the scene's background) is executed as soon as it accumulated enough evidence, suggesting that this object should be explored in more detail.
Integrating visual information in an object-based manner is motivated by recent work in perceptual psychology, that relates human eye movement behavior and object-based target selection. 
A clear definition of an \textit{object}, however, is challenging and highly depends upon the task. In this work, humans, animals and vehicles are defined as objects, whereas the remaining portion of the scene is assigned to a general background category.

By incorporating recent successes in computer vision, ranging from low-level saliency predictions to semantic object segmentation, and integrating new findings of perceptual psychology, our model is capable of \textit{implicitly} generating natural human viewing behavior. Aspects of human eye movements such as foveation duration, smooth pursuit, and saccade characteristics are \textit{not explicitly} implemented, but they arise as a consequence of our object-based decision making process.


\section{Eye movements as a sequential decision making task}
\label{sec:model}
In this section, we introduce a new model for object-based saccadic decision making called ObjectDDM, shown in Fig.~\ref{fig:title}.
A feature map $F$ is computed for each video frame based on bottom-up features. 
Here, we choose a low-level saliency map  for $F$, based on motion, intensity, color and orientation \cite{molin2015motion}, multiplied with a Gaussian anisotropic center bias (as suggested in \cite{nuthmann2017well}). 
The visual sensitivity map $S$ depends on the current gaze position $(x_0,y_0)$ and determines  the extent to which features can be perceived by the observer. We approximate this sensitivity with an isotropic Gaussian $G = \frac{1}{2\pi\sigma^2}\exp\left(-\frac{(x-x_0)^2+(y-y_0)^2}{2\sigma^2}\right)$, to account for higher acuity in the fovea compared to the periphery. 
Based on psychophysical evidence for an object-based spread of covert attention, the  part of the Gaussian distribution that falls within the currently foveated object is replaced with uniform sensitivity $\Omega$.
To compute the evidence $\mu_i$ for an eye movement towards the respective object, the visual information $(F\cdot S)$ is then masked by the individual object segmentation mask\footnote{We extract object segmentation masks on a frame-by-frame basis using MaskRCNN \cite{he2017mask} and track segmented objects using DeepSort \cite{wojke2017simple}}  $O_i$.  We adapt the common practice to scale the perceptual object size logarithmically (cf.\ \cite{nuthmann2017well}).
Consequently, visual evidence for each object $O_i$ present in the scene is computed as follows,
\begin{eqnarray}
    \mu_i = \frac{\sum_{x=1}^{N} O_i^x \cdot (S^x \cdot F^x)}{\sum_{x=1}^{N} O_i^x} \cdot  \log\sum\limits_{x=1}^{N} O_i^x
\end{eqnarray}
where $x$ is a particular pixel location and $N$ the total number of pixels in the video frame. 

Based on the evidence for each object, the respective decision variable $V_i$ is then updated with,
\begin{equation*}
dV_i = \begin{cases}
\mu_i \cdot dt + s \cdot dW &\text{if foveated}\\
\mu_i \cdot (1 - \xi_i) \cdot dt + s \cdot dW &\text{else}
\end{cases}
\end{equation*}
where $s$ determines the amount of diffusion and $dW \sim \mathcal{N}(0,1)$. 
The inhibition of return $\xi$ to an object (IOR) is set to one if it is foveated, and linearly decreases with rate $\epsilon$ towards zero if it is not foveated: $d\xi = -\epsilon dt$.
This mechanism enables exploration of novel locations. 
If for one object the decision variable $V_i$ reaches the decision threshold $\theta$, a saccade is triggered towards this target. 
The exact landing position within the object is probabilistic, and proportional to the information $(S\cdot F)$ at each location. 

Across saccades, the predicted gaze location moves with the currently foveated object mask, resulting in fixation or smooth pursuit behavior depending on the object motion.

\section{Experiments}
\label{sec:experiments}
\textbf{Dataset: \ }
We explore and evaluate our model on high resolution videos of natural scenes from the VidCom dataset \cite{li2011visual}. This data comes with high-quality human eye-movement data (14 subjects on 50 videos with 300 frames) from a free-viewing task, providing ground truth. We used the Deep EM classifier \cite{startsev20191d} for gaze classification in saccade and foveation (combination of fixation and smooth pursuit) events.
Although our approach can in principle handle an arbitrary number of objects, for practical reasons we excluded videos if they are too crowded ($>10$ objects) or if the object tracking algorithm failed. This would introduce challenges for our drift-diffusion model since evidence for an object could not be accumulated reliably over time. The remaining 23 clips were randomly split in 8 videos for parameter exploration and 15 for evaluation.
Based on the parameter exploration on the training set we define an operating point for our model.  
The results of this parametrization for the previously unseen test videos is shown in the following.

\textbf{Evaluation metrics: \ }
Due to the high dimensionality and variability of scanpaths, there is currently no universal agreement on how to evaluate the quality of a predicted scanpath from a computational model. 
We set out multiple evaluation benchmarks to asses the validity of predicted scanpaths based on their global statistics.
First and foremost, a well-performing model should generate a realistic number of saccades in each video, and reproduce the distribution of foveation durations between them. 
We further evaluated the distributions of saccade amplitudes and directions.
In order to asses how realistic the scene exploration behavior of the predicted scanpaths is, we assessed the ratio of saccades that are made within the same object\footnote{Saccades made within the perceptual background are not counted towards within-object saccades.}. 
Additionally, and as a proxy for visual working memory, we evaluate the average time until a previously foveated object is re-fixated. 


\textbf{Baseline model: \ }
First steps towards scanpath prediction for dynamic scenes have been made by Fuhl et al.\ \cite{fuhl2018eye} who simulate eye movements primarily for the evaluation of eye-tracking software.
A scanpath in their model is based on a randomly generated sequence of eye movements containing saccades and fixations (no smooth pursuit), where the event duration is sampled from a predefined (uniform) distribution within a specific range. 
This sequence is then mapped on a video by using the frame-wise local maxima of three saliency maps as potential fixation targets. 


\begin{figure*}
    \centering
    \includegraphics[width=0.94\textwidth]{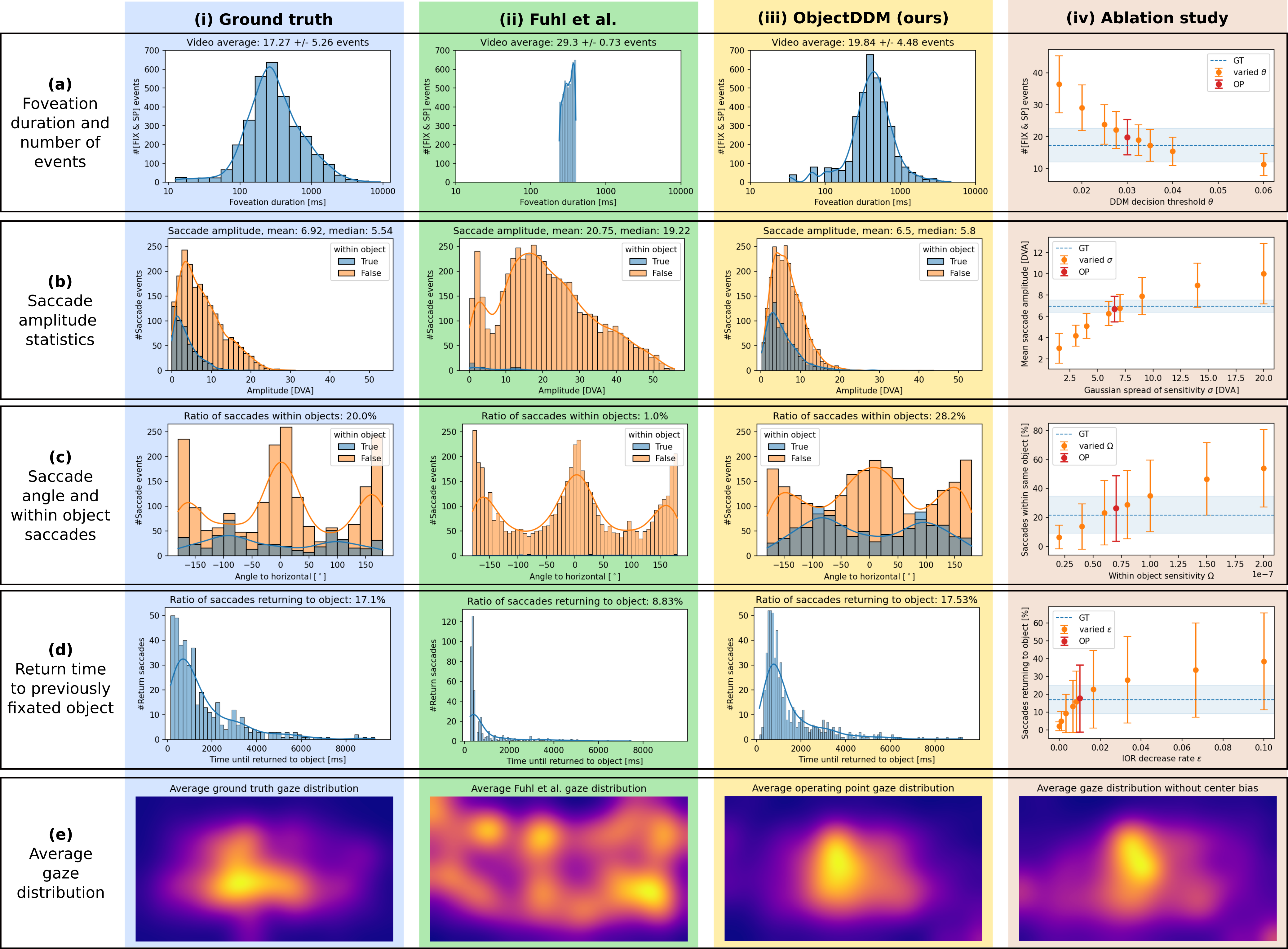}
    \caption{Overview of experiments and evaluation metrics. (i) Scanpath statistics of 14 subjects (on average 12 per video). (ii) Saccade generator by Fuhl et al.\ \cite{fuhl2018eye} with default parameters. (iii) Our object-based drift-diffusion model (parameter ``operating point'' (OP) based on training set: $\theta=0.03, s=0.002, \sigma=6.5\text{ DVA}, \Omega=7\cdot10^{-8}, \epsilon=1/120$). Both models are run for 12 random seeds per video. (iv) We vary individual parameters (others fixed to values in (iii)) to show their specific influence on evaluation metrics. We plot the mean value and standard deviation of the respective measure across the videos. 
    }
    \label{fig:experiments}
\end{figure*}

\textbf{Scanpath prediction in dynamic scenes: \ }
    Figure~\ref{fig:experiments} shows a comparison between human ground truth scanpath statistics with estimates from the scanpath generator by Fuhl et al.\ \cite{fuhl2018eye} and our here proposed ObjectDDM.     
    Free parameters in our model are: the DDM decision threshold $\theta$ and noise level $s$, the visual sensitivity spread around the gaze point as Gaussian with $\sigma$ and constant value  $\Omega$ within the fixated object, and the linear decrease of the IOR $\epsilon$ (see section \ref{sec:model}). 
    Their values are fixed based on a parameter exploration on the training set and we demonstrate their effect on the evaluation metrics on the test data in an ablation study.
    It is important to note that the observed distributions from our model are not predefined, but arise as a consequence of our scene-aware decision process based on biologically plausible mechanisms.
    
    Fixation durations in the ground-truth data approximate a lognormal distribution (see Fig.~\ref{fig:experiments}(a)). This shape is reproduced by our model as a direct result of how we accumulate evidence in the DDM. Fuhl et al., on the other hand, specify a uniform distribution to describe fixation times, which does not fit the experimental data. 
    The number of saccadic decisions triggered by the DDM mainly depends on the decision threshold $\theta$. The DDM noise level $s$ 
    leads to a higher variance between different runs of the model. 

    The general shape of the amplitude distribution, with a fast decrease towards larger amplitudes and many small saccades within objects, is captured by the ObjectDDM (cf.\ Fig.~\ref{fig:experiments}(b)); saccades with very small or big amplitudes are however underrepresented. A higher standard deviation $\sigma$ of the Gaussian aperture 
    leads to more frequent saccades to distant objects, as well as to larger saccades towards (or within) the background. 
    If saccade targets are exclusively based on peaks in the saliency map, as in Fuhl et al., this results in an unrealistic amplitude distribution. 
    
    The saccade-direction distribution is known to have characteristic peaks in the cardinal directions ($0^\circ, 90^\circ, 180^\circ, -90^\circ$, see Fig.~\ref{fig:experiments}(c)). The bias towards horizontal saccades can be observed in both Fuhl et al.\ and ObjectDDM.
    As shown in the ground truth data, vertical saccades occur disproportionately often within objects, due to their vertical extension. This bias is also reproduced in our results.
    About every fifth saccade in the ground truth is made to further investigate the currently fixated object. Our results show a higher ratio in total (28\%),  but are similar to human behavior across most videos (see Fig.~\ref{fig:experiments}(c-iv)). This ratio strongly depends on the sensitivity $\Omega$ for the currently fixated object. Higher values result in an increasing number of saccades within the same object and consequently a smaller number of large exploratory saccades. 
    The object agnostic gaze generator by Fuhl et al.\ shows almost no saccades within objects (only 1\%).
    
    Humans use about one in six saccades to re-foveate an object that they already explored previously. The time course of this exploration statistic is shown in Fig.~\ref{fig:experiments}(d).
    The results from the ObjectDDM match the ratio of re-foveation events in the ground truth and their decrease over time. We however observe a lack of immediate return saccades, suggesting that the assumption of full inhibition of an object after foveation is overly simplified. 
    The importance of the IOR mechanism is demonstrated by varying its decrease rate $\epsilon$. Large values correspond to short inhibition times, often resulting in fast re-foveations of the most salient objects. Long lasting inhibitions (small $\epsilon$), on the other hand, force the model to ``ignore'' previously foveated objects. 

    In Fig.~\ref{fig:experiments}(e) we observe the well known phenomenon that, on average, people predominantly look at the center of the screen. This bias is partially explained by a tendency to look straight ahead, but also due to a clustering of interesting targets in the center of the scene. While the scanpath generator -- based on low-level saliency without object notions -- does not show this bias, it is clearly visible in the ObjectDDM results. 
    By excluding the Gaussian center bias from the calculation of the feature maps, we can asses its importance for our model predictions.
    The direct comparison between Fig.~\ref{fig:experiments}(e-iii) and (e-iv) shows that the tendency to look at (and pursue) objects already describes the average gaze distribution well; adding a center bias is not in fact required.

\section{Conclusion}
In the proposed modelling framework, scanpaths are generated for dynamic real-world scenes as a result of an object-based, noisy accumulation process of visual information. 
To the best of our knowledge, this is the first model that predicts the spatial and temporal aspects of eye movements for dynamic scenes. 
We show that a small number of parameters suffices to reproduce important statistics of previously unseen human eye-tracking data.
This framework can easily be modified or extended towards additional attention mechanisms, different feature maps, or alternative object representations. 
The effect of individual mechanisms of attentional control can be explored (and quantified) using the ObjectDDM model by varying the corresponding model parameters.
Consequently, we hope that this work will motivate new hypothesis and experimental work on attention allocation in dynamic real-world scenes.


\

\section*{Acknowledgment}
Funded by the German Research Foundation under Germany’s Excellence Strategy – EXC 2002/1 “Science of Intelligence” – project number 390523135.
{\small

\bibliographystyle{IEEEtran}
\bibliography{egbib}
}

\end{document}